\documentclass{article}

\usepackage{microtype}
\usepackage{enumitem}
\usepackage{siunitx}
\usepackage{graphicx}
\usepackage{subfigure}
\usepackage{amsmath}
\usepackage{amssymb}
\usepackage{booktabs} % for professional tables
\usepackage{algorithm}
\usepackage{hyperref}

\usepackage[accepted]{icml2018}

\icmltitlerunning{Variational Bayesian Approximation to Log Determinants}

\begin{document}

\twocolumn[
\icmltitle{VBALD - Variational Bayesian Approximation of Log Determinants}



\icmlsetsymbol{equal}{*}

\begin{icmlauthorlist}
\icmlauthor{Diego Granziol}{goo,to}
\icmlauthor{Stephen Roberts}{to,goo}
\icmlauthor{Michael Osborne}{goo, to}
\end{icmlauthorlist}

\icmlaffiliation{to}{University of Oxford}
\icmlaffiliation{goo}{Oxford Man Institute}

\icmlkeywords{Machine Learning, ICML}

\vskip 0.3in
]



\printAffiliationsAndNotice{\icmlEqualContribution} % otherwise use the standard text.

\begin{abstract}
Evaluating the log  determinant  of  a positive  definite  matrix  is  ubiquitous  in  machine learning. Applications thereof range from Gaussian processes, minimum-volume ellipsoids, metric learning, kernel learning, Bayesian neural networks, Determinental Point Processes, Markov random fields to partition functions of discrete graphical models. In order to avoid the canonical, yet prohibitive, Cholesky $\mathcal{O}(n^{3})$ computational cost, we propose a novel approach, with complexity $\mathcal{O}(n^{2})$, based on a constrained variational Bayes algorithm. We compare our method to Taylor, Chebyshev and Lanczos approaches and show state of the art performance on both synthetic and real-world datasets.
\end{abstract}

\section{Introduction}
Algorithmic scalability is a keystone in the realm of modern machine learning. Making high quality inference, on large, feature rich datasets under a constrained computational budget is arguably the primary goal of the learning community. A common hindrance, appearing in Gaussian graphical models, Gaussian Processes \cite{rue2005gaussian,rasmussen2006gaussian}, sampling, variational inference \cite{mackay2003information}, metric/kernel learning \cite{davis2007information,van2009minimum}, Markov random fields \cite{wainwright2006log}, Determinantal Point Processes (DPP's) and Bayesian Neural networks \cite{mackay1992bayesian}, is the calculation of the log determinant of a large positive definite matrix. 

For a large positive definite matrix $K \in  \mathcal{R}^{n\times n}$, the canonical solution involves the Cholesky decomposition, $K = LL^{T}$. The log determinant is then trivial to calculate as $\log \mathrm{Det}(K) = 2\sum_{i=1}^{n}\log L_{ii}$. This computation invokes a computational complexity $\mathcal{O}(n^{3})$ and storage complexity $\mathcal{O}(n^{2})$ and is thus unfit for purpose for $n>10^{4}$, i.e. even a small sample set in the age of big data. 

\subsection{Related Work}
Recent work in machine learning combined stochastic trace estimation with Taylor approximations for Gaussian process parameter learning \cite{zhang2007approximate, boutsidis2017randomized}, reducing the computational cost to matrix vector multiplications (MVMs), $\mathcal{O}(n^{2})$ for a dense matrix and $\mathcal{O}(nnz)$\footnote{Number of non zeros.} for a sparse matrix. Variants of the same theme used Chebyshev polynomial approximations, giving a performance improvement over Taylor along with refined error bounds \cite{han2015large} and the combination of either Chebyshev/Lanczos techniques with structured kernel interpolation (SKI) in order to accelerate MVMs to $\mathcal{O}(n+i\log i)$, where $i$ is the number of inducing points \cite{dong2017scalable}. 

This approach relies on an extension of Kronecker and Toeplitz methods, which are limited to low dimensional (typically $D \leq 5$) data, which cannot be assumed in general.  Secondly, whilst Lanczos methods have a convergence rate of double that of the Chebyshev approaches, the derived bounds require $\mathcal{O}(\sqrt{\kappa})$ Lanczos steps \cite{Ubaru2016}, where $\kappa$ is the matrix condition number. In many practical cases of interest $\kappa > 10^{10}$ and thus the large number of $m$ matrix vector multiplications becomes prohibitive. %Furthermore we require the storage of the Lancsoz vectors which must be repeatedly re-orthogonalised. This brings the total computational complexity to $\mathcal{O}(nnz(K)+nm^{2})$, which can majorly reduce the speed benefits for sparse matrices.
We restrict ourselves to the high-dimensional, high-condition number, big data limit.

\subsection{Contribution}
We recast the problem of calculating log determinants as a constrained variational inference problem, which under the assumption of a uniform prior reduces to maximum entropy spectral density estimation given moment information. We develop a novel algorithm using Newton conjugate gradient and Hessian information with a Legendre/Chebyshev basis, as opposed to power moments. Our algorithm is stable for a large number of moments ($m > 30$) surpassing the $m \approx 8$ limit of previous MaxEnt algorithms \cite{DBLP:conf/bigdataconf/GranziolR17,bandyopadhyay2005maximum,mead1984maximum}. We build on the experimental results of \cite{han2015large,bild}, adding to the case against using Taylor expansions in Log Determinant approximations by proving that the implied density violates Kolmogorov's axioms of probability.  We compare our algorithm to both Chebyshev and Lanczos methods on real data and synthetic kernel matrices of arbitrary condition numbers, relevant to GP kernel learning and DPP's.

The work most similar to ours is \cite{ete}, which uses stochastic trace estimation as moment constraints with an off the shelf maximum entropy algorithm \cite{bandyopadhyay2005maximum} to estimate the matrix spectral density in order to calculate the log determinant. Their approach outperforms Chebyshev, Taylor, Lanczos and kernel based approximations, yet begins to show pathologies and increasing error for  $m \geq 8$ moments, due to convergence issues \cite{DBLP:conf/bigdataconf/GranziolR17}, making it unsuitable for machine learning where typical squared exponential kernels have very sharply decaying spectra and thus a larger number of moments is required for high precision. 
\section{Motivating example}
Determinantal point processes (DPPs) \cite{macchi1975coincidence} are probabilistic models capturing global negative correlations. They describe Fermions \footnote{as a consequence of the spin-statistics theorem} in Quantum Physics, Eigenvalues of random matrices and non-intersecting random walks. 

In machine learning their natural selection of diversity has found applications in the field of summarization \cite{gong2014diverse}, human pose detection \cite{kulesza_2012}, clustering \cite{kang2013fast}, Low rank kernel matrix approximations \cite{li2016fast} and Manifold learning \cite{wachinger2015sampling}.

Formally, it defines a distribution on $2^{y}$, where $y = [n]$ is the finite ground set. For a random variable $X \subseteq Y$ drawn from a given DPP we have
\begin{equation}
P(X = x) \propto \mathrm{det}(K_{X}) = \frac{\mathrm{det}(K_{x})}{\mathrm{det}(K+I)},
\end{equation}
where $K \in \mathbb{R}^{d\times d}$ is a positive definite matrix referred to as the $L$-ensemble kernel. Greedy algorithms that find the most diverse set $Y$ of $y$ that achieves the highest probability, i.e $\mathrm{argmax}_{X\subseteq y} \mathrm{det}(K_{Y})$ require the calculation of the marginal gain, 
\begin{equation}
\log \mathrm{det} K_{X\cup\{i\}} - \log \mathrm{det} L_{X}.
\end{equation}
with $\mathcal{O}(n^{3})$ computational complexity. Previous work has looked at limiting the burden of the computational complexity by employed Chebyshev approximations to the Log Determinant \cite{han2017faster}. However their work is limited Kernel Matrices with minimum eigenvalues of $10^{-3}$\footnote{or alternatively low condition numbers}, which does not cover the class of realistic kernel matrix spectra, notably the popular squared exponential kernel. We develop a method in section \ref*{method} and an algorithm \ref{algorithm} which is capable of handling very high condition numbered matrices.

\section{Background}
\subsection{Log Determinants as a Density Estimation Problem}

Any symmetric positive definite (PD) matrix $K$, is diagonalizable by a unitary transformation $U$, i.e $K = U^{t}DU$, where $D$ is the matrix with the eigenvalues of $K$ along the diagonal. Hence we can write the log determinant as:
\begin{equation}
\log \text{Det} K = \log \prod_{i}\lambda_{i} = \sum_{i=1}^{n}\log\lambda_{i} = n\mathbb{E}_{\mu}(\log \lambda).
\end{equation} 
Here we have used the cyclicity of the determinant and $\mathbb{E}_{\mu}$ denotes the expectation under the spectral measure. The latter can be written as:
\begin{equation}
\label{eq:spectrallog}
\begin{aligned}
\mathbb{E}_{\mu}(\log \lambda) & =  \int_{\lambda_{min}}^{\lambda_{max}}d\mu(\lambda)\log \lambda  \\  & = \int_{\lambda_{min}}^{\lambda_{max}}\sum_{i=1}^{n}\frac{1}{n}\delta(\lambda-\lambda_{i})\log \lambda d\lambda.
\end{aligned}
\end{equation}
Given that the matrix is PD, we know that $\lambda_{min}>0$ and we can divide the matrix by an upper bound, $\lambda_{u} \geq \lambda_{max}$, via the Gershgorin circle theorem \cite{gershgorin1931uber} such that,
\begin{equation}
\begin{aligned}
& \log \text{Det} \frac{K}{\lambda_{u}} =  n\mathbb{E}_{\mu}(\log \lambda') =  n\mathbb{E}_{\mu}(\log \lambda) -  n\lambda_{u}\\
& \therefore \log \text{Det} K = n\mathbb{E}_{\mu}(\log \lambda') + n\lambda_{u}.
\end{aligned}
\end{equation}
Here $\lambda_{u} = \text{arg max}_{i}(\sum_{j=1}^{n}|K_{ij}|)$, i.e the max sum of the rows of the absolute of the matrix $K$. Hence we can comfortably work with the transformed measure,
\begin{equation}
\int_{\lambda_{min}/\lambda_{u}}^{\lambda_{max}/\lambda_{u}}p(\lambda')\log \lambda' d\lambda' = \int_{0}^{1}p(\lambda')\log \lambda' d\lambda',
\end{equation}
as the spectral density $p(\lambda)$ is $0$ outside of its bounds, which are bounded by $[0,1]$ respectively.

\subsection{Stochastic Trace Estimation}
Using the expectation of quadratic forms, for any multivariate random variable $v$ with mean $m$ and variance $\Sigma$, we can write 
\begin{equation}
\mathbb{E}(zz^{t}) = mm^{t}+\Sigma \xrightarrow[m = 0]{\Sigma = I} I,
\end{equation}
where in the last equality we have assumed that the variable possesses zero mean and unit variance. By the linearity of trace and expectation for any $m\geq 0$ we can write
\begin{equation}
\sum_{i=1}^{n}\lambda^{m} = n\mathbb{E}_{\mu}(\lambda^{m}) = \text{Tr}(IK^{m}) = \mathbb{E}(zK^{m}z^{t}).
\end{equation}
In practice we approximate the expectation over all random vectors with a simple Monte Carlo average. i.e for $d$ random vectors ,
\begin{equation}
 \mathbb{E}(zK^{m}z^{t}) \approx \frac{1}{d}\bigg(\sum_{j=1}^{j=d}z_{j}K^{m}z_{j}^{t} \bigg),
\end{equation}
where we take the product of the matrix $K$ with the vector $z_{j}$, $m$ times, so as to avoid costly $\mathcal{O}(n^{3})$ matrix matrix multiplication. This allows us to calculate the non central moment expectations in $\mathcal{O}(dmn^{2})$ for dense matrices, or $\mathcal{O}(dm\times nnz)$ for sparse matrices, where $d\times m << n$.

The random unit vector $z_{j}$ can be drawn from any distribution, such as a Gaussian. Choosing the components of $z_{j}$ to be i.i.d Rademacher random variables i.e $P(+1)=P(-1) = \frac{1}{2}$ (Hutchinson's method \cite{hutchinson1990stochastic}) has the lowest variance of such estimators \cite{jackmub}, satisfying,
\begin{equation}
\text{Var}[\text{Tr}K] = 2\bigg(||K||^{2}-\sum_{i=1}^{n}K_{ii}^{2}\bigg).
\end{equation}
Loose bounds exist on the number of samples $d$ required to get within a fractional error $\epsilon$ with probability $1-\eta$ \cite{han2015large},
\begin{equation}
\begin{aligned}
& d \geq  6\epsilon^{-2}\log(\frac{2}{\eta})\\
& \text{Pr}\bigg[\frac{|\text{Tr}_{est}(K)-\text{Tr}(K)|}{\text{Tr}(K)}  \leq \epsilon\bigg] \geq 1-\eta,
\end{aligned}
\end{equation}
as per \cite{roosta2015improved}. To get within $1\%$ fractional error with probability $0.9$, for example, we require $d = 180,000$ samples. In practice we find that as little as $d = 30$ gives good accuracy however.

\section{Polynomial approximations to the Log Determinant}
Recent work \cite{han2015large,dong2017scalable,zhang2007approximate} has considered incorporating knowledge of the non central moments \footnote{Also using stochastic trace estimation.} of a normalised eigenspectrum by replacing the logarithm with a finite polynomial expansion,
\begin{equation}
\label{polynomialexpansion}
\mathbb{E}_{\mu}  =  \int_{0}^{1}p(\lambda)\log (\lambda) d\lambda = \int_{0}^{1}p(\lambda)\log (1-(1-\lambda)) d\lambda.
\end{equation}
Given that $\log(\lambda)$ is not analytic at $\lambda = 0$, it can be seen that, for any density with a large mass near the origin, a very large number of polynomial expansions, and thus moment estimates, will be required to achieve a good approximation, irrespective of the choice of basis.

\subsection{Taylor approximations are probabilistically unsound}
\label{taylorisshit}
In the case of a Taylor expansion equation \eqref{polynomialexpansion} can be written as,
\begin{equation}
-\int_{0}^{1} p(\lambda)\sum_{i=1}^{\infty}\frac{(1-\lambda)^{i}}{i} \approx  -\int_{0}^{1} p(\lambda)\sum_{i=1}^{n}\frac{(1-\lambda)^{i}}{i}.
\end{equation}
The error in this approximation can be written as the difference of the two sums,
\begin{equation}
\label{taylorerror}
-\sum_{i=m+1}^{n}\frac{\mathbb{E}_{\mu}(1-\lambda)^{i}}{i},
\end{equation}
where we have used the Taylor expansion of $\log(1-x)$ and $\mathbb{E}_{\mu}$ denotes the expectation under the spectral measure.

De-Finetti \cite{finetti}  showed that Kolmogorov's axioms of probability \cite{kolmogorov_1950} could be derived by manipulating probabilities in such a manner so as to not make a sure loss on a gambling system based on them. Such a probabilistic framework, of which the Bayesian is a special case \cite{bayesianspecialcase}, satisfies the conditions of,
\begin{enumerate}
	\item \bfseries{Non Negativity:} $p_{i} \geq 0 \thinspace \forall i$,
	\item Normalization: $\sum_{i}p_{i} = 1$,
	\item Finite Additivity: $P(\cup_{n=1}^{N}A_{n}) = \sum_{n=1}^{N}P(A_{n})$.~~~\footnote{for a sequence of disjoint sets $A_{n}$.}
\end{enumerate}
The intuitive appeal of De-Finetti's sure loss arguments, is that they are inherently performance based. A sure loss is a practical cost, which we wish to eliminate. 

Keeping within such a very general formulation of probability and thus inference. We begin with complete ignorance about the spectral density $p(\lambda)$ (other than its domain $[0,1]$) and by some scheme after seeing the first $m$ non-central moment estimates we propose a surrogate density $q(\lambda)$. The error in our approximation can be written as,
\begin{eqnarray}
\label{spectralerror}
\int_{0}^{1} [p(\lambda)-q(\lambda)]\log(\lambda)d\lambda \nonumber \\
 = \int_{0}^{1} -[p(\lambda)-q(\lambda)]\sum_{i=1}^{\infty}\frac{(1-\lambda)^{i}}{i}d\lambda.
\end{eqnarray}
For this error to be equal to that of our Taylor expansion \eqref{taylorerror}, our implicit surrogate density must have the first $m$ non-central moments of $(1-\lambda)$ identical to the true spectral density $p(\lambda)$ and all others $0$. %Given that we know that the spectral density $p(\lambda)$ is a sum of $n$ Dirac delta distributions with unknown locations\footnote
	%{We could of course do $n$ sets of non central moment estimates using stochastic trace estimation and then solve for the location of the dirac delta functions exactly, however for a dense matrix this operation is $\mathcal{O}(n^{3})$ and hence no better than using the Cholesky decomposition so we are back where we started} bounded by the region $[0,1]$ the only case for which this is possible, would be for the $n$ delta distributions to co-locate at $\lambda=1$, This corresponds to the identity matrix, whose log determinant is trivially $0$ and the spectral expectation $E_{\mu}(1-\lambda)^{i} = 0,\thinspace \forall i$.

For any PD matrix $K$, for which $E_{\mu}(1-\lambda)^{i} > 0,\thinspace \forall i\leq m$\footnote{we except the trivial case of a Dirac distribution at $\lambda=1$, which is of no practical interest}, for equation \eqref{spectralerror} to be equal to \eqref{taylorerror}, we must have,
\begin{equation}
\int_{0}^{1}q(\lambda)\sum_{m+1}^{\infty}\frac{(1-\lambda)^{i}}{i}d\lambda = 0.
\end{equation}
Given that $ 0 \leq \lambda \leq 1$ and that we have removed the trivial case of the spectral density (and by implication its surrogate) being a delta function at $\lambda = 1$, the sum is manifestly positive and hence $q(\lambda) < 0$ for some $\lambda$, which is incompatible with the theory of probability \cite{finetti,kolmogorov_1950}.

\section{Constrained Variational Method}
\subsection{Variational Bayes}
Variational Methods \cite{mackay2003information,fox2012tutorial} in machine learning pose the problem of intractable density estimation from the application of Bayes' rule as a functional optimization problem,
\begin{equation}
p(z|x) = \frac{p(x|z)p(z)}{p(x)} \approx q(z),
\end{equation}
and finding finding the appropriate $q(z)$. 
 
Typically, whilst the functional form of $p(x|z)$ is known, calculating $p(x) = \int p(x|z)p(z)dz$ is intractable. Using Jensen's inequality we can show that,
\begin{equation}
\log p(x) \geq \mathbb{E}_{q}[\log p(x,z)]-\mathbb{E}_{q}[\log q(z)]. \footnote{Where the R.H.S is known as the evidence lower bound (ELBO).}
\end{equation}
 
It can be shown that the reverse KL divergence between the posterior and the variational distribution, $\mathbb{D}_{kl}(q|p)$, can be written as, 
\begin{equation}
\log p(x) = \mathbb{E}_{q}[\log p(x,z)]-\mathbb{E}_{q}[\log q(z)] + \mathbb{D}_{kl}(q|p).
\end{equation}
Hence maximising the evidence lower bound is equivalent to minimising the reverse KL divergence between $p$ and $q$. 

By assuming the variational distribution to factor over the set of latent variables, the functional form of the variational marginals, subject to the normalisation condition, can be computed using functional differentiation of the dual,
\begin{equation}
\frac{\partial}{\partial Q_{i}(x_{i})}\bigg\{-\mathbb{D}_{kl}[Q_{i}(x_{i}|Q^{*}(x_{i}))]-\lambda_{i}\bigg(\int Q_{i}dx_{i}-1\bigg)\bigg\},
\end{equation}
leading to a Gibbs' distribution and an iterative update equation. 

\subsection{Log Determinant as a Variational Inference problem}
\label{method}
We consider minimizing the reverse KL divergence between our surrogate posterior $q(\lambda)$ and our prior $p_{0}(\lambda)$ on the eigenspectrum, 
\begin{equation}
\mathcal{D}_{kl}(q|p_{0}) = -H(q) - \int_{0}^{1} q(\lambda)\log p_{0}(\lambda)d\lambda,
\end{equation}
such that the normalization and moment constraints are satisfied. Here $H(q)$ denotes the differential entropy of the density $q$.

By the theory of Lagrangian duality, the convexity of the KL divergence and the affine nature of the moment constraints, we maximise the dual \cite{boyd_vandenberghe_2009},
\begin{equation}
-H(q) - \int q(\lambda)\log p_{0}(\lambda)d\lambda - \sum_{i=0}^{m}\alpha_{i}\biggl(\int_{0}^{1}q(\lambda)\lambda^{i}d\lambda - \mu_{i}\biggr),
\end{equation}
or alternatively we minimise
\begin{equation}
\label{eq:maxrelent}
H(q) + \int q(\lambda)\log p_{0}(\lambda)d\lambda - \sum_{i=0}^{m}\alpha_{i}\biggl(\int_{0}^{1}q(\lambda)\lambda^{i}d\lambda - \mu_{i}\biggr).
\end{equation}

\subsection{Link to Information Physics}
In the field of information physics the minimization of Equation \eqref{eq:maxrelent} is known as the method of relative entropy \cite{caticha2012entropic}. It can be derived as the unique functional satisfying the axioms of,
\begin{enumerate}
	\item {\bfseries Locality:} local information has local effects.
	\item {\bfseries Co-ordinate invariance:} the co-ordinate set carries no information.
	\item {\bfseries Sub-System Independence:} for two independent sub-system it should not matter if we treat the inference separately or jointly.
	\item {\bfseries Objectivity:} Under no new information, our inference should not change. Hence under no constraints, our posterior should coincide with our prior.
\end{enumerate}
These lead to the generalised entropic functional,
\begin{equation}
-\int q(x)\log\frac{q(x)}{m(x)}dx - \sum_{i}\alpha_{i}\bigg(\int_{x\in \mathcal{D}}f_{i}(x)dx-\mu_{i}\bigg).
\end{equation}
Here the justification for restricting ourselves to a functional is derived from considering the set of all distributions $q_{i}(\lambda)$ compatible with the constraints and devising a transitive ranking scheme.  It can be shown, further, that Newton's laws, non-relativistic quantum mechanics and Bayes' rule can all be derived under this formalism.  

In the case of a flat prior over the spectral domain, we reduce to the method of maximum entropy with moment constraints \cite{jaynes1982rationale,inftheoryjaynes}. Conditions for the existence of a solution to this problem have been proved for the case of the Hausdorff moment conditions \cite{mead1984maximum}, of which our problem is a special case. 
\subsection{Algorithm}
\label{algorithm}
The generalised dual objective function which we minimise is,
\begin{equation}
\mathcal{S}(q,q_{0}) = \int_{0}^{1}q_{0}(\lambda)\exp(-[1+\sum_{i}\alpha_{i}\lambda^{i}])d\lambda + \sum_{i}\alpha_{i}\mu_{i},
\end{equation}
which can be shown to have gradient
\begin{equation}
\frac{\partial \mathcal{S}(q,q_{0})}{\partial \alpha_{j}}= \mu_{j}-\int_{0}^{1}q_{0}(\lambda)\lambda^{j}\exp(-[1+\sum_{i}\alpha_{i}\lambda^{i}])d\lambda,
\end{equation}
and Hessian
\begin{equation}
\frac{\partial^{2} \mathcal{S}(q,q_{0})}{\partial \alpha_{j}\partial\alpha_{k}}= \int_{0}^{1}q_{0}(\lambda)\lambda^{j+k}\exp(-[1+\sum_{i}\alpha_{i}\lambda^{i}])d\lambda.
\end{equation}

\subsubsection{Prior Spectral Belief}
If we assume complete ignorance over the spectral domain, then the natural maximally entropic prior is the uniform distribution and hence $q(\lambda) = 1$.\footnote{Technically as the log determinant exists and is finite, we cannot have any mass at $\lambda=0$, hence we must set the uniform between some $[\delta\epsilon,1]$, where $\delta\epsilon >0$.}
An alternative prior over the $[0,1]$ domain is the Beta distribution, the maximum entropy distribution of that domain under a mean and log mean constraint,
\begin{equation}
\frac{\Gamma(\gamma+\beta)}{\Gamma(\gamma)\Gamma(\beta)}\lambda^{\gamma-1}(1-\lambda)^{\beta-1}.
\end{equation}
The log mean constraint is particularly interesting as we know that it must exists for a valid log determinant to exist, as is seen for equation \eqref{eq:spectrallog}.  We set the parameters of by maximum likelihood, hence,
\begin{equation}
\gamma = \frac{\mu_{1}(\mu_{1}-\mu_{2})}{\mu_{2}-\mu_{1}^{2}}\thinspace , \thinspace \beta = \bigg(\frac{1}{\mu_{1}}-1\bigg)\frac{\mu_{1}(\mu_{1}-\mu_{2})}{\mu_{2}-\mu_{1}^{2}}\thinspace.
\end{equation}

\subsubsection{Analytical surrogate form}
Our final equation for $q(\lambda)$ can be written as,
\begin{equation}
q(\lambda) = \frac{\Gamma(\gamma+\beta)}{\Gamma(\gamma)\Gamma(\beta)}\lambda^{\gamma-1}(1-\lambda)^{\beta-1}\times\exp(-[1+\sum_{i=0}^{m}\alpha_{i}\lambda^{i}])
\end{equation}
for the beta prior and 
\begin{equation}
q(\lambda) = \exp(-[1+\sum_{i=0}^{m}\alpha_{i}\lambda^{i}])
\end{equation}
for the uniform. The exponential factor can be thought of altering the prior beta/uniform distribution so as to fit the observed moment information. 
\subsubsection{Practical Implementation}
For simplicity we have kept all the formula's in terms of power moments, however we find vastly improved performance and conditioning when we switch to another polynomial basis. Many alternative and orthogonal Polynomial bases exist (so that the errors between moment estimations are uncorrelated), we implement both Chebyshev and Legendre moments in our Lagrangian and find similar performance for both. The use of Chebyshev moments in Machine Learning and Computer Vision has been reported to be of practical siginifcance previously \cite{yap2001chebyshev}. We use Python's SciPy minimize standard newton-conjugate gradient algorithm to solve the objective, given the gradient and hessian to within a gradient tolerance $gtol$. To make the Hessian better conditioned so as to achieve convergence we add jitter along the diagonal. The pseudo code is given in Algorithm \ref{alg:example}. The Log Determinant is then calculated using Algorithm \ref{alg:logdet}.

\begin{algorithm}[tb]
	\caption{VBALD}
	\label{alg:example}
	
	\begin{algorithmic}[1]
		\STATE {\bfseries Input:} Moments $\{\mu_{i}\}$, Tolerance $\epsilon$, Hessian noise $\eta$
		\STATE {\bfseries Output:} Coefficients $\{\alpha_{i}\}$
		\STATE {\bfseries Do:} Newton-CG 
		\STATE Initialize $\alpha_{i} = 0$.
		\STATE Minimize $\int_{0}^{1}q_{0}(\lambda)q_{c}(\lambda)d\lambda + \sum_{i}\alpha_{i}\mu_{i}$
		\STATE Gradient $\mu_{j}-\int_{0}^{1}q_{0}(\lambda)q_{c}(\lambda)\lambda^{j}d\lambda$
		\STATE H  $ = \int_{0}^{1}q_{0}(\lambda)q_{c}(\lambda)\lambda^{j+k}d\lambda$
		\STATE Hessian $= (H+H')/2 + \eta$
		\STATE {\bfseries Until:} gtol $< \epsilon$
		
		%\UNTIL{$noChange$ is $true$}
	\end{algorithmic}
\end{algorithm}

\begin{algorithm}
	\caption{Computing Log Determinant using Constrained Variational Inference}\label{alg:logdet}
	\begin{algorithmic}[1]
		\vspace{0.5em}
		\STATE {\bfseries Input:} PD Symmetric Matrix $A$, Order of stochastic trace estimation $k$, Tolerance $\epsilon$
		\STATE {\bfseries Output:} Log Determinant Approximation $\log|K|$
		\STATE $B = K/\lambda_{u}$
		\STATE $\mu$ (moments)$ \gets$ StochasticTraceEstimation$(B, k)$ 
		\STATE $\alpha$ (coefficients) $\gets \text{VBALD(}\mu, \epsilon)$
		\STATE $q(\lambda) \gets q(\lambda | \alpha)$
		\STATE $\log|A| \gets n\int \log(\lambda) q(\lambda) d\lambda + n\log(\lambda_{u})$
	\end{algorithmic}
\end{algorithm}

\section{Experiments}
In order to test the validity and practical applicability of our proposed Algorithm, we test on both Synthetic and real Kernel matrices. We compare against Chebyshev and Lanczos approaches. For completeness in Figures \ref{fig:lscale0.1},\ref{fig:lscale0.33} and \ref{fig:lscale0.66} we include results from the Taylor expansion. However in light of the consistently superior performance of the Chebyshev approach \cite{han2015large}, the arguments in Section \ref{taylorisshit} and space requirements we do not include Taylor methods in our results tables.

\subsection{Synthetic Kernel Data}
\begin{figure}
	\caption{Comparison of VBALD against Taylor, Lanczos and Chebyshev algorithms, absolute relative error on the $y$-axis and number of moments used on the $x$-axis.}
	\begin{subfigure}%{width=0.2\textwidth}
		\centering
		\includegraphics[width=.8\linewidth]{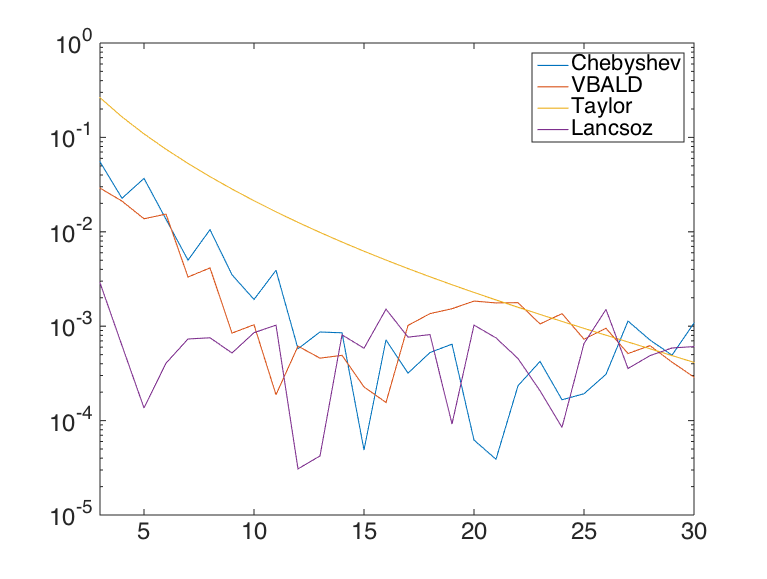}
		\caption{Length Scale = 0.1, Condition number = 16}
		\label{fig:lscale0.1}
	\end{subfigure}
	\begin{subfigure}%{.2\textwidth}
		\centering
		\includegraphics[width=.8\linewidth]{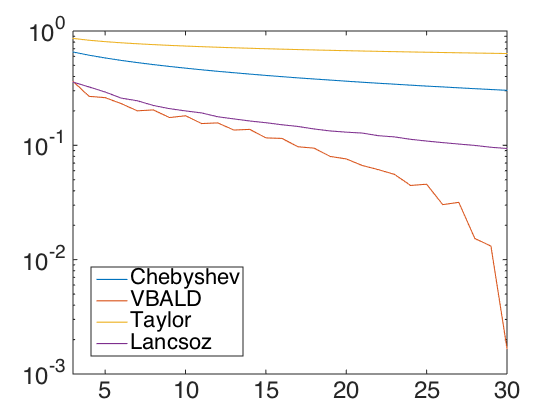}
		\caption{Length Scale = 0.33, Condition number = $2\times 10^{7}$ Equivalent Chebyshev steps $n=1200$, Lanczos steps $n\approx 100$.}
		\label{fig:lscale0.33}
	\end{subfigure}
	\begin{subfigure}%{.2\textwidth}
		\centering
		\includegraphics[width=.8\linewidth]{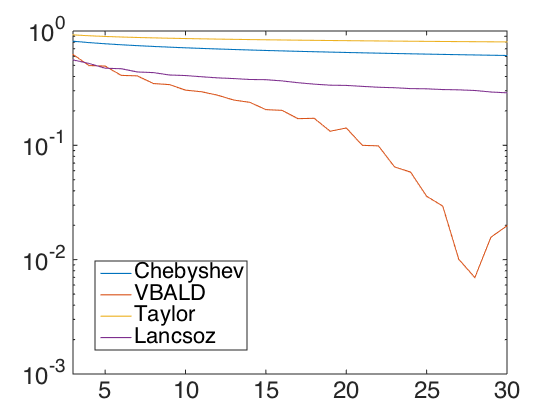}
		\caption{Length Scale = 0.66, Condition number = $1.8 \times 10^{11}$ equivalent Chebyshev steps $n>20000$  Lanczos steps $n\approx 500$.}
		\label{fig:lscale0.66}
	\end{subfigure}
\end{figure}
We simulate the kernel matrices from a Gaussian/Determinental Point Process \cite{Rasmussen2006}, by generating a typical squared exponential kernel matrix $K \in \mathbb{R}^{n \times n}$ using the Python GPy package with $6$ dimensional, Gaussian inputs. We then add noise of variance $10^{-8}$ along the diagonals. We employ a variety of realistic uniform length-scales ($0.1$, $0.33$ \& $0.66$) with condition numbers $16$, $2\times 10^{7}$  \& $1.8\times 10^{11}$ respectively. We use $m=30$ Moments, $d=50$ Hutchinson probe vectors and compare VBALD against the Taylor approximation, Chebyshev \cite{han2015large} and Lanczos \cite{ubaru2017fast}. We see that for low condition numbers (Figure \ref{fig:lscale0.1}) The benefit of framing the log determinant as an optimization problem is marginal, whereas for large condition numbers (Figures \ref{fig:lscale0.33} \& \ref{fig:lscale0.66}) the benefits are substantial, with orders of magnitude better results than competing methods. We also provide the number of Chebyshev and Lanczos steps required to achieve similar performance. Results for a wider range of length scales are shown in Table \ref{table1}. 
\begin{table}[t]
	\label{table1}
	\caption{Absolute relative error for VBALD, Chebyshev \& Lanczos methodson varying length-scale $l$, with varying condition number $\kappa$ on squared exponential kernel matrices $K \in \mathbb{R}^{1000\times 1000}$.}
	\label{sample-table}
	\vskip 0.15in
	\begin{center}
		\begin{small}
			\begin{sc}
				\begin{tabular}{lcccr}
					\toprule
					$\kappa$ & $l$ & VBALD & Chebyshev & Lanczos \\
					\midrule
					$3\times10^{1}$    & 0.05& \bf{0.0014}& 0.0037 & 0.0024 \\
					$1.1\times10^{3}$ & 0.15& 0.0522& \bf{0.0104} & 0.0024\\
					$1.0\times10^{5}$    & 0.25& 0.0387& 0.0795 & \bf{0.0072} \\
					$2.4\times10^{6}$    & 0.35& 0.0263& 0.2302    & \bf{0.0196}    \\
					$8.3\times10^{7}$     & 0.45& \bf{0.0284}& 0.3439 & 0.0502\\
					$4.2\times10^{8}$     & 0.55&
					\bf{0.0256}& 0.4089 & 0.0646\\
					$4.3\times10^{9}$     & 0.65&
					\bf{0.00048}& 0.5049 & 0.0838\\
					$1.4\times10^{10}$     & 0.75& \bf{0.0086}& 0.5049  &0.1050       \\
					$4.2\times10^{10}$   & 0.85& \bf{0.0177}& 0.5358 &0.1199\\
					\bottomrule
				\end{tabular}
			\end{sc}
		\end{small}
	\end{center}
	\vskip -0.1in
\end{table}
\subsection{Sparse Matrices}
Given that the method described above works in general for any positive definite matrices. We look at large sparse spectral problems. Given that good results are reported for Polynomial methods at even a fairly modest number of moments, we consider very very large sparse matrices, such as social networks\footnote{Facebook, with $10^{9}$ users, each with $\approx 10^{3}$ friends is equivalent to a dense $10^{6}$ matrix}, where we wish to limit the number of non-central moment estimates. We set $m=d=5$ and test on the UFL SuiteSparse dataset. We show our results in table \ref{table2}. VBALD it competitive with Lanczos and we note that similar to the Kernel Matrix case, VBALD does best relative to other methods when the others perform at their poorest, i.e have a relatively large absolute relative error.
\begin{table}[t]
	\label{table2}
	\caption{Absolute relative error for VBALD, Chebyshev \& Lanczos on SuiteSparse Matrices}
	\label{sample-table}
	\vskip 0.15in
	\begin{center}
		\begin{small}
			\begin{sc}
				\begin{tabular}{lcccr}
					\toprule
					Dataset & $m$ & VBALD & Chebyshev & Lanczos \\
					\midrule
					Thermo    & 5 & $1\times 10^{-2}$& $5\times 10^{-2}$ & $\bf{3\times 10^{-3}}$ \\
					Water 2    & 5& $4\times 10^{-3}$& $1\times 10^{-2}$ & $\bf{9\times 10^{-4}}$ \\
					Water 1    & 5& $\bf{2\times 10^{-4}}$& $3\times 10^{-3}$ & $2\times 10^{-4}$ \\
					jnlbrng1    & 5& $3\times 10^{-2}$& $3\times 10^{-2}$ & $\bf{2\times 10^{-2}}$ \\
					finan512    & 5& $9\times 10^{-3}$& $8\times 10^{-2}$ & $\bf{1\times 10^{-3}}$ \\
					
					Ecology 2    & 5& $\bf{1\times 10^{-2}}$& $1\times 10^{-2}$ & $3\times 10^{-2}$ \\
					Apache    & 5& $\bf{5\times 10^{-2}}$& \num{2e-1}    & \num{8e-2}    \\

					\bottomrule
				\end{tabular}
			\end{sc}
		\end{small}
	\end{center}
	\vskip -0.1in
\end{table}

% Note use of \abovespace and \belowspace to get reasonable spacing
% above and below tabular lines.

\subsection{Citations and References}

\subsection{Software and Data}

\section*{Acknowledgements}


\bibliography{sample}

\begin{thebibliography}{39}
\providecommand{\natexlab}[1]{#1}
\providecommand{\url}[1]{\texttt{#1}}
\expandafter\ifx\csname urlstyle\endcsname\relax
  \providecommand{\doi}[1]{doi: #1}\else
  \providecommand{\doi}{doi: \begingroup \urlstyle{rm}\Url}\fi

\bibitem[Bandyopadhyay et~al.(2005)Bandyopadhyay, Bhattacharya, Biswas, and
  Drabold]{bandyopadhyay2005maximum}
Bandyopadhyay, K, Bhattacharya, Arun~K, Biswas, Parthapratim, and Drabold, DA.
\newblock Maximum entropy and the problem of moments: A stable algorithm.
\newblock \emph{Physical Review E}, 71\penalty0 (5):\penalty0 057701, 2005.

\bibitem[Boutsidis et~al.(2017)Boutsidis, Drineas, Kambadur, Kontopoulou, and
  Zouzias]{boutsidis2017randomized}
Boutsidis, Christos, Drineas, Petros, Kambadur, Prabhanjan, Kontopoulou,
  Eugenia-Maria, and Zouzias, Anastasios.
\newblock A randomized algorithm for approximating the log determinant of a
  symmetric positive definite matrix.
\newblock \emph{Linear Algebra and its Applications}, 533:\penalty0 95--117,
  2017.

\bibitem[Boyd \& Vandenberghe(2009)Boyd and
  Vandenberghe]{boyd_vandenberghe_2009}
Boyd, Stephen~P. and Vandenberghe, Lieven.
\newblock \emph{Convex optimization}.
\newblock Cambridge University Press, 2009.

\bibitem[Caticha(2012)]{caticha2012entropic}
Caticha, A.
\newblock Entropic inference and the foundations of physics (monograph
  commissioned by the 11th brazilian meeting on {B}ayesian
  statistics--ebeb-2012, 2012.

\bibitem[Davis et~al.(2007)Davis, Kulis, Jain, Sra, and
  Dhillon]{davis2007information}
Davis, Jason~V, Kulis, Brian, Jain, Prateek, Sra, Suvrit, and Dhillon,
  Inderjit~S.
\newblock Information-theoretic metric learning.
\newblock In \emph{Proceedings of the 24th international conference on Machine
  learning}, pp.\  209--216. ACM, 2007.

\bibitem[De~Finetti(1974)]{finetti}
De~Finetti, Bruno.
\newblock Teoria delle probabilita. einaudi, turin, 1970.
\newblock \emph{English translation:[51]}, 1974.

\bibitem[Dong et~al.(2017)Dong, Eriksson, Nickisch, Bindel, and
  Wilson]{dong2017scalable}
Dong, Kun, Eriksson, David, Nickisch, Hannes, Bindel, David, and Wilson,
  Andrew~G.
\newblock Scalable log determinants for gaussian process kernel learning.
\newblock In \emph{Advances in Neural Information Processing Systems}, pp.\
  6330--6340, 2017.

\bibitem[Fitzsimons et~al.(2016)Fitzsimons, Osborne, Roberts, and
  Fitzsimons]{jackmub}
Fitzsimons, J.~K., Osborne, M.~A., Roberts, S.~J., and Fitzsimons, J.~F.
\newblock Improved stochastic trace estimation using mutually unbiased bases,
  2016.

\bibitem[Fitzsimons et~al.(2017{\natexlab{a}})Fitzsimons, Cutajar, Osborne,
  Roberts, and Filippone]{bild}
Fitzsimons, Jack, Cutajar, Kurt, Osborne, Michael, Roberts, Stephen, and
  Filippone, Maurizio.
\newblock Bayesian inference of log determinants, 2017{\natexlab{a}}.

\bibitem[Fitzsimons et~al.(2017{\natexlab{b}})Fitzsimons, Granziol, Cutajar,
  Osborne, Filippone, and Roberts]{ete}
Fitzsimons, Jack, Granziol, Diego, Cutajar, Kurt, Osborne, Michael, Filippone,
  Maurizio, and Roberts, Stephen.
\newblock Entropic trace estimates for log determinants, 2017{\natexlab{b}}.

\bibitem[Fox \& Roberts(2012)Fox and Roberts]{fox2012tutorial}
Fox, Charles~W and Roberts, Stephen~J.
\newblock A tutorial on variational bayesian inference.
\newblock \emph{Artificial intelligence review}, 38\penalty0 (2):\penalty0
  85--95, 2012.

\bibitem[Gershgorin(1931)]{gershgorin1931uber}
Gershgorin, Semyon~Aranovich.
\newblock Uber die abgrenzung der eigenwerte einer matrix.
\newblock \emph{Известия Российской академии
  наук. Серия математическая}, \penalty0 (6):\penalty0
  749--754, 1931.

\bibitem[Gong et~al.(2014)Gong, Chao, Grauman, and Sha]{gong2014diverse}
Gong, Boqing, Chao, Wei-Lun, Grauman, Kristen, and Sha, Fei.
\newblock Diverse sequential subset selection for supervised video
  summarization.
\newblock In \emph{Advances in Neural Information Processing Systems}, pp.\
  2069--2077, 2014.

\bibitem[Granziol \& Roberts(2017)Granziol and
  Roberts]{DBLP:conf/bigdataconf/GranziolR17}
Granziol, Diego and Roberts, Stephen~J.
\newblock Entropic determinants of massive matrices.
\newblock In \emph{2017 {IEEE} International Conference on Big Data, BigData
  2017, Boston, MA, USA, December 11-14, 2017}, pp.\  88--93, 2017.
\newblock \doi{10.1109/BigData.2017.8257915}.
\newblock URL \url{https://doi.org/10.1109/BigData.2017.8257915}.

\bibitem[Han et~al.(2015)Han, Malioutov, and Shin]{han2015large}
Han, Insu, Malioutov, Dmitry, and Shin, Jinwoo.
\newblock Large-scale log-determinant computation through stochastic chebyshev
  expansions.
\newblock In \emph{International Conference on Machine Learning}, pp.\
  908--917, 2015.

\bibitem[Han et~al.(2017)Han, Kambadur, Park, and Shin]{han2017faster}
Han, Insu, Kambadur, Prabhanjan, Park, Kyoungsoo, and Shin, Jinwoo.
\newblock Faster greedy map inference for determinantal point processes.
\newblock \emph{arXiv preprint arXiv:1703.03389}, 2017.

\bibitem[Hutchinson(1990)]{hutchinson1990stochastic}
Hutchinson, Michael~F.
\newblock A stochastic estimator of the trace of the influence matrix for
  laplacian smoothing splines.
\newblock \emph{Communications in Statistics-Simulation and Computation},
  19\penalty0 (2):\penalty0 433--450, 1990.

\bibitem[Jaynes(1957)]{inftheoryjaynes}
Jaynes, E.~T.
\newblock Information theory and statistical mechanics.
\newblock \emph{Phys. Rev.}, 106:\penalty0 620--630, May 1957.
\newblock \doi{10.1103/PhysRev.106.620}.
\newblock URL \url{http://link.aps.org/doi/10.1103/PhysRev.106.620}.

\bibitem[Jaynes(1982)]{jaynes1982rationale}
Jaynes, Edwin~T.
\newblock On the rationale of maximum-entropy methods.
\newblock \emph{Proceedings of the IEEE}, 70\penalty0 (9):\penalty0 939--952,
  1982.

\bibitem[Kang(2013)]{kang2013fast}
Kang, Byungkon.
\newblock Fast determinantal point process sampling with application to
  clustering.
\newblock In \emph{Advances in Neural Information Processing Systems}, pp.\
  2319--2327, 2013.

\bibitem[Kolmogorov(1950)]{kolmogorov_1950}
Kolmogorov, A.~N.
\newblock On logical foundations of probability theory.
\newblock \emph{Lecture Notes in Mathematics Probability Theory and
  Mathematical Statistics}, pp.\  1–5, 1950.
\newblock \doi{10.1007/bfb0072897}.

\bibitem[Kulesza(2012)]{kulesza_2012}
Kulesza, Alex.
\newblock Determinantal point processes for machine learning.
\newblock \emph{Foundations and Trends® in Machine Learning}, 5\penalty0
  (2-3):\penalty0 123–286, 2012.
\newblock \doi{10.1561/2200000044}.

\bibitem[Li et~al.(2016)Li, Jegelka, and Sra]{li2016fast}
Li, Chengtao, Jegelka, Stefanie, and Sra, Suvrit.
\newblock Fast dpp sampling for nystr$\backslash$" om with application to
  kernel methods.
\newblock \emph{arXiv preprint arXiv:1603.06052}, 2016.

\bibitem[Macchi(1975)]{macchi1975coincidence}
Macchi, Odile.
\newblock The coincidence approach to stochastic point processes.
\newblock \emph{Advances in Applied Probability}, 7\penalty0 (1):\penalty0
  83--122, 1975.

\bibitem[MacKay(1992)]{mackay1992bayesian}
MacKay, David~JC.
\newblock \emph{Bayesian methods for adaptive models}.
\newblock PhD thesis, California Institute of Technology, 1992.

\bibitem[MacKay(2003)]{mackay2003information}
MacKay, David~JC.
\newblock \emph{Information theory, inference and learning algorithms}.
\newblock Cambridge university press, 2003.

\bibitem[Mead \& Papanicolaou(1984)Mead and Papanicolaou]{mead1984maximum}
Mead, Lawrence~R and Papanicolaou, Nikos.
\newblock Maximum entropy in the problem of moments.
\newblock \emph{Journal of Mathematical Physics}, 25\penalty0 (8):\penalty0
  2404--2417, 1984.

\bibitem[Rasmussen \& Williams(2006)Rasmussen and Williams]{Rasmussen2006}
Rasmussen, Carl~E. and Williams, Christopher.
\newblock \emph{{Gaussian Processes for Machine Learning}}.
\newblock MIT Press, 2006.

\bibitem[Rasmussen(2006)]{rasmussen2006gaussian}
Rasmussen, Carl~Edward.
\newblock Gaussian processes for machine learning.
\newblock 2006.

\bibitem[Roosta-Khorasani \& Ascher(2015)Roosta-Khorasani and
  Ascher]{roosta2015improved}
Roosta-Khorasani, Farbod and Ascher, Uri.
\newblock Improved bounds on sample size for implicit matrix trace estimators.
\newblock \emph{Foundations of Computational Mathematics}, 15\penalty0
  (5):\penalty0 1187--1212, 2015.

\bibitem[Rue \& Held(2005)Rue and Held]{rue2005gaussian}
Rue, Havard and Held, Leonhard.
\newblock \emph{Gaussian Markov random fields: theory and applications}.
\newblock CRC press, 2005.

\bibitem[Ubaru et~al.(2016)Ubaru, Chen, and Saad]{Ubaru2016}
Ubaru, Shashanka, Chen, Jie, and Saad, Yousef.
\newblock Fast {E}stimation of tr (f (a)) via {S}tochastic {L}anczos
  {Q}uadrature.
\newblock 2016.

\bibitem[Ubaru et~al.(2017)Ubaru, Chen, and Saad]{ubaru2017fast}
Ubaru, Shashanka, Chen, Jie, and Saad, Yousef.
\newblock Fast estimation of tr(f(a)) via stochastic lanczos quadrature.
\newblock \emph{SIAM Journal on Matrix Analysis and Applications}, 38\penalty0
  (4):\penalty0 1075--1099, 2017.

\bibitem[Van~Aelst \& Rousseeuw(2009)Van~Aelst and Rousseeuw]{van2009minimum}
Van~Aelst, Stefan and Rousseeuw, Peter.
\newblock Minimum volume ellipsoid.
\newblock \emph{Wiley Interdisciplinary Reviews: Computational Statistics},
  1\penalty0 (1):\penalty0 71--82, 2009.

\bibitem[Wachinger \& Golland(2015)Wachinger and
  Golland]{wachinger2015sampling}
Wachinger, Christian and Golland, Polina.
\newblock Sampling from determinantal point processes for scalable manifold
  learning.
\newblock In \emph{International Conference on Information Processing in
  Medical Imaging}, pp.\  687--698. Springer, 2015.

\bibitem[Wainwright \& Jordan(2006)Wainwright and Jordan]{wainwright2006log}
Wainwright, Martin~J and Jordan, Michael~I.
\newblock Log-determinant relaxation for approximate inference in discrete
  markov random fields.
\newblock \emph{IEEE transactions on signal processing}, 54\penalty0
  (6):\penalty0 2099--2109, 2006.

\bibitem[Walley(1991)]{bayesianspecialcase}
Walley, Peter.
\newblock Statistical reasoning with imprecise probabilities.
\newblock 1991.
\newblock \doi{10.1007/978-1-4899-3472-7}.

\bibitem[Yap et~al.(2001)Yap, Raveendran, and Ong]{yap2001chebyshev}
Yap, PT, Raveendran, P, and Ong, SH.
\newblock Chebyshev moments as a new set of moments for image reconstruction.
\newblock In \emph{Neural Networks, 2001. Proceedings. IJCNN'01. International
  Joint Conference on}, volume~4, pp.\  2856--2860. IEEE, 2001.

\bibitem[Zhang \& Leithead(2007)Zhang and Leithead]{zhang2007approximate}
Zhang, Yunong and Leithead, William~E.
\newblock Approximate implementation of the logarithm of the matrix determinant
  in gaussian process regression.
\newblock \emph{Journal of Statistical Computation and Simulation}, 77\penalty0
  (4):\penalty0 329--348, 2007.

\end{thebibliography}
\bibliographystyle{icml2018}

\appendix
\section{Do \emph{not} have an appendix here}

\textbf{\emph{Do not put content after the references.}}
Put anything that you might normally include after the references in a separate
supplementary file.

We recommend that you build supplementary material in a separate document.
If you must create one PDF and cut it up, please be careful to use a tool that
doesn't alter the margins, and that doesn't aggressively rewrite the PDF file.
pdftk usually works fine. 

\textbf{Please do not use Apple's preview to cut off supplementary material.} In
previous years it has altered margins, and created headaches at the camera-ready
stage. 
%%%%%%%%%%%%%%%%%%%%%%%%%%%%%%%%%%%%%%%%%%%%%%%%%%%%%%%%%%%%%%%%%%%%%%%%%%%%%%%
%%%%%%%%%%%%%%%%%%%%%%%%%%%%%%%%%%%%%%%%%%%%%%%%%%%%%%%%%%%%%%%%%%%%%%%%%%%%%%%

\end{document}